\documentclass{article}
\usepackage{spconf,amsmath,graphicx}

\usepackage{enumitem}
\setlist{nosep, leftmargin=14pt}

\usepackage{mwe} 
\usepackage{comment}

\title{A Novel Autoencoders-LSTM Model for Stroke Outcome Prediction using Multimodal MRI Data}

 \name{Nima Hatami$^{1}$ $\quad$ Laura Mechtouff$^{2,3}$ $\quad$ David Rousseau$^{4}$ $\quad$ Tae-Hee Cho$^{2,3}$ $\quad$  Omer Eker$^{1,2}$ $\quad$\\
 \em Yves Berthezène$^{1,2}$ $\qquad$ Carole Frindel$^{1}$ }

\name{Nima Hatami$^{1}$ $\quad$ Laura Mechtouff$^{2,3}$ $\quad$ David Rousseau$^{4}$ $\quad$ Tae-Hee Cho$^{2,3}$ $\quad$  Carole Frindel$^{1}$ }

\address{\normalsize $^{1}$CREATIS, CNRS UMR5220, INSERM U1206, Université Lyon 1, INSA-Lyon, France\\
        \normalsize $^{2}$Stroke Department, Hospices Civils de Lyon, France\\
        \normalsize $^{3}$ CarMeN, INSERM U1060, INRA U1397, Université Lyon 1, INSA-Lyon, France\\
        \normalsize $^{4}$LARIS, UMR IRHS INRA, Universite d’Angers, France
        }
        
\begin{document}
%

\maketitle{}

\begin{abstract}

Patient outcome prediction is critical in management of ischemic stroke. In this paper, a novel machine learning model is proposed for stroke outcome prediction using multimodal Magnetic Resonance Imaging (MRI). The proposed model consists of two serial levels of Autoencoders (AEs), where different AEs at level 1 are used for learning unimodal features from different MRI modalities and a AE at level 2 is used to combine the unimodal features into compressed multimodal features. The sequences of multimodal features of a given patient are then used by an LSTM network for predicting outcome score. The proposed AE$^2$-LSTM model is proved to be an effective approach for better addressing the multimodality and volumetric nature of MRI data. 
Experimental results show that the proposed AE$^2$-LSTM outperforms the existing state-of-the art models by achieving highest AUC=0.71 and lowest MAE=0.34.

\end{abstract}
\section{Introduction}
\label{sec:intro}

Stroke is the second-leading cause of death and the third-leading cause of death and disability combined \cite{feigin2021global}. 
About 87\% of all strokes are classified as ischemic. In this case, an artery that supplies blood to the brain is blocked. 
Although reperfusion therapies (intravenous thrombolysis and thrombectomy) have revolutionized ischemic stroke management, outcome remains highly heterogeneous across patients. Improving the prediction of outcome in
stroke patients would contribute to tailor treatment decisions and evaluate novel therapeutic strategies.

Application of machine learning models on stroke outcome prediction is a field of growing interest. Some authors proposed different 2D and 3D Convolutional Neural Networks (CNNs) for predicting patient outcomes from MR/CT images and clinical data \cite{ramos2020predicting,bacchi2020deep,zihni2020,nishi2020deep,hatami2022EMBC,hatami2022ESOC}. However, none of them perfectly match the multimodal and volumetric nature of MRI/CT data or contain too many parameters (in case of 3D-CNNs) when the number of image modalities increases. 

A two-level Autoencoders followed by a Long Short-Term Memory (AE$^2$-LSTM) is proposed for stroke outcome prediction using multimodal MRI data. Five AEs are first trained to learn unimodal features from five MRI modalities. Another AE is used to combine these five unimodal features into one main multimodal representation. And finally, an LSTM network is applied to predict the 3-month modified Rankin Scale (mRS), a 7-point disability scale, from the compressed multimodal feature series. 

The rest of the paper is organized as follows: a review of recent machine learning approaches on stroke outcome prediction is presented in the next section. The proposed AE$^2$-LSTM framework is described in detail in section 3. A description of the dataset used in this study and experimental results are reported in section 4. Finally, section 5 concludes the paper and outlines the future directions.

\section{Related Work}
\label{sec:related}

\begin{figure*}[htb]
\begin{center}
\includegraphics[scale=0.32]{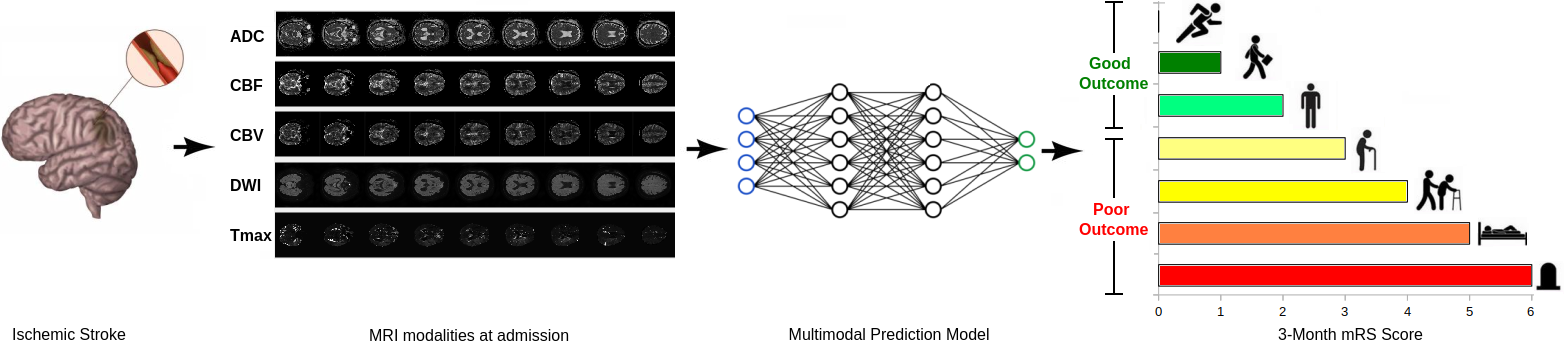}
\end{center}
\vspace{-0.2cm}
\caption{The general block diagram of a machine learning model for stroke outcome prediction using multimodal MRI.}
\label{fig1}
\end{figure*}


The general block diagram of a machine learning-based stroke outcome prediction using multimodal MRI is shown in Figure \ref{fig1}. 
In one of the early efforts to predict patient outcome using machine learning, Ramos et al. \cite{ramos2020predicting} applied Random Forest (RF), Support Vector Machine (SVM) and Artficial Neural Networks (ANN) algorithms to predict mRS scores. Both clinical/biological variables -- e.g. age, pre-stroke mRS, glucose level and NIH Stroke Scale (NIHSS) at baseline -- and radiological parameters -- presence of leukoaraiosis, old infarctions, hyperdense vessel sign, and hemorrhagic transformation -- were used as input features. The main limitation of such models is that the features are hand-crafted, unlike the state-of the art end-to-end deep learning models that learned dynamically and automatically features from raw images. 

Bacchi et al. \cite{bacchi2020deep} proposed to combine CNN and ANN models to predict NIHSS or mRS scores. Noncontrast CT images are processed via a CNN model while clinical data (such as age, sex, blood pressure) are processed in an ANN model. The outputs of both models are then merged through Fully-Connected (FC) layers to generate the final binary (poor vs. good) outcomes.

Zihni et al. \cite{zihni2020} proposed a deep learning-based multimodal fusion strategy for integration of neuroimaging information with clinical metadata. First, a 3D-CNN and a ANN models are built for processing neuroimaging and clinical data, respectively. Then, features from two models are fused using FC layer for prediction of mRS scores. 

Nishi et al. \cite{nishi2020deep} applied deep learning for extracting neuroimaging features in order to predict clinical outcomes (mRS) for patients with large vessel occlusion. They proposed a 2-output deep encoder-decoder architecture (3D U-Net) with DWI data as an input. The U-Net prediction mask (output 1) is for the ischemic lesion segmentation task, and the high-level feature maps (output 2) are also extracted from the deepest middle layers. Then, they used the obtained features in a 2-layer ANN to predict the binarized 3-month mRS scores.

Recently, Hatami et al. \cite{hatami2022ESOC} proposed a multimodal CNN-LSTM model to automatically encode the spatio-temporal context of MR images in a deep learning architecture. They applied a curriculum learning, a family of learning algorithms in which first starts with only "easy" patients (from the model's perspective) and then gradually includes more "difficult" patients. The training has two main steps: first a CNN-LSTM ranks the patients from 1 to 10 based on their "difficulty", and then another CNN-LSTM (similar architecture) uses the ranking information to progressively learn the mRS scores by including different types of patients ranked according to their difficulty at different training epochs.

A multimodal CNN-LSTM based ensemble model is proposed in \cite{hatami2022EMBC} for processing five MR imaging modalities. For each modality, a dedicated network provides preliminary prediction of the mRS. The final mRS score is obtained by merging the preliminary probabilities of each module dedicated to a specific type of MR image weighted by a clinical metadata, i.e. age or NIHSS. 

The main limitation of the previous models is that none of them perfectly fits the multimodal and volumetric nature of the MRI data. For instance, 3D CNNs in \cite{bacchi2020deep,zihni2020,nishi2020deep} result in too many parameters if they use a dedicated 3D-CNN for each image modality. Also, CNN-LSTM models proposed by \cite{hatami2022EMBC,hatami2022ESOC} use CNNs pre-trained on natural image dataset (ImageNet \cite{ImageNet15}). 
The authors used ImageNet features of off-the-shelf models such as ResNet \cite{ResNet}, DenseNet \cite{DenseNet} and VGG \cite{VGG16} with no fine-tuning on MRI data which is not ideal. The proposed AE$^2$-LSTM Model tries to address these challenges. 

\section{Method}
\label{sec:method}

\begin{figure*}[htb]
\begin{center}
\includegraphics[scale=0.3]{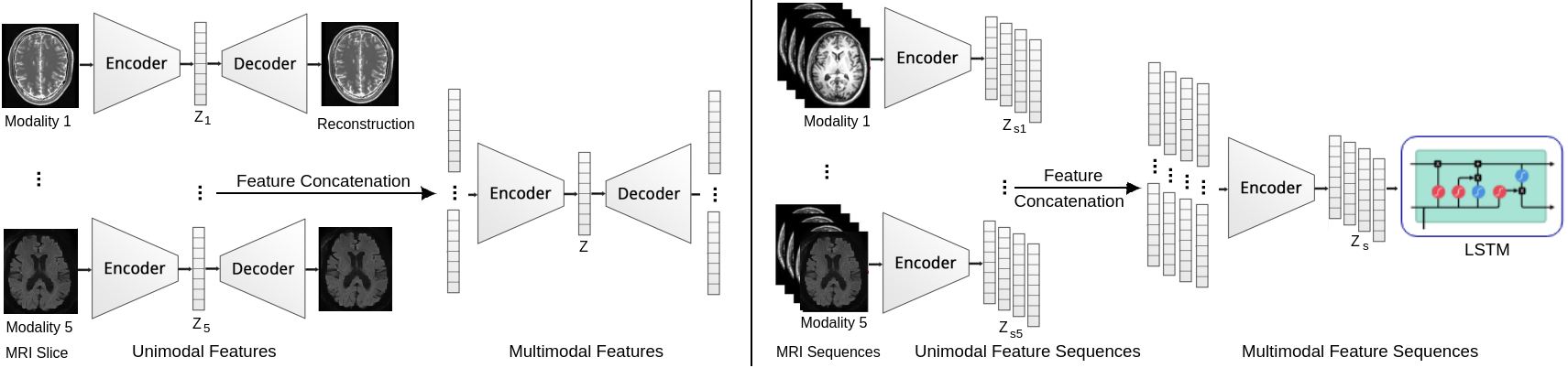}
\end{center}
\caption{The proposed AE$^2$-LSTM model for stroke outcome prediction using multimodal MRI. Left: slice-level training of AEs for learning unimodal and multimodal features. Right: patient-level training and testing of LSTM for predicting mRS score.}
\label{fig3}
\end{figure*}

\subsection{Autoencoders (AE)}
An AE is a special type of neural network with an unsupervised learning technique for the task of representation learning. It is trained to attempt to copy its input to its output. The network consists of two parts: an encoder function $h=f(x)$ and a decoder that produces a reconstruction $r=g(h)$. Blocks of AEs can be seen in Figure \ref{fig3} left.
The AE learns a representation (encoding) for a set of data, typically for dimensionality reduction, by training the network to ignore insignificant information (“noise”). Different variations of AEs are proposed aiming to force the learned representations to assume useful properties \cite{goodfellow2016deep}. For example regularized AEs (Sparse, Denoising and Contractive) are proposed for learning representations (features) for classification tasks \cite{vincent2010stacked} and Variational AEs are used as generative models \cite{kingma2019introduction}.

\subsection{Long Short Term Memory Networks (LSTM)}

LSTM \cite{LSTM97} is a type of recurrent neural network to classify, process and predict time series data. Unlike standard feedforward neural networks, LSTM has feedback connections. A typical LSTM unit is composed of a cell, an input gate, an output gate and a forget gate. The cell remembers values over arbitrary time intervals and the gates regulate the flow of information into and out of the cell. A block of the LSTM network can be seen in Figure \ref{fig3} right.

\subsection{Proposed AE$^2$-LSTM Model}

The proposed AE$^2$-LSTM model for stroke outcome prediction using multimodal MRI data is shown in Figure \ref{fig3}. This novel architecture is composed of two levels of AEs (an AE at level 1 is specific to each of the imaging modality given as input, and level 2 AE is used to combine the unimodal representations into multimodal features) followed by an LSTM network. It is proposed to encode the volumetric nature of MRI sequences into a series of compressed features, each feature vector representing an MRI slice. Each patient is therefore represented by a series of multimodal feature series, which is later used by an LSTM network to map them into the patient outcome. The training phase has two main steps: 

i) training of AEs for learning unimodal and multimodal features. In this step which is shown in Figure \ref{fig3} left, there are five AEs at level 1, one for each MRI modality, and one AE at level 2 for combining them. AEs at level 1 are responsible for learning unimodal features ($Z_{1-5}$) from MRI slices. These feature vectors are compressed representations of each modality. The AE at level 2 is receiving concatenated $Z_{1-5}$ features as input and output, and trying to compress it further into the final multimodal features ($Z$). This is the vector representing each slice from five modalities. This step can be considered as unsupervised multimodal feature learning of MRI sequences. It is important to note that the AEs can be trained in two manners: AEs viewed isolated and independent from each other and minimize the reconstruction loss functions separately (which is the case in this paper), or all AEs being considered as one big model and minimize the losses simultaneously (e.g. minimizing sum of the losses).

ii) training of LSTM for predicting mRS score. In this step which is shown in Figure \ref{fig3} right, a series of multimodal features for each patient are generated by inferring the AE encoders. Afterwards, the LSTM network is trained in order to map each sequential feature into a final mRS score. 

In the testing phase and for each given patient, AE encoders are applied to the MRI sequences in order to obtain the feature series $Z_s$, and the LSTM network is used to predict the final mRS score (Figure \ref{fig3} right).

In this research, we used five MR imaging modalities, i.e. Apparent Diffusion Coefficient (ADC), Cerebral Blood Flow (CBF), Cerebral Blood Volume (CBV),  Diffusion-Weighted Imaging (DWI) and Time-to-Maximum (Tmax). However, the model is flexible, and any number of modalities or other types of imaging (e.g. CT) can be used. 

\section{Experiments}
\label{sec:experiments}

\begin{table*}[t]
\caption{Performance (mean $\pm$ std over 10 runs) of the proposed AE$^2$-LSTM compared to different baselines.}


\begin{center}
\begin{tabular}{lllllll}
\begin{tabular}{c c c c c c c c}
\hline
Model & Input data & AUC  & MAE & Accuracy &  Specificity  &  Sensitivity  & F1-score \\[2pt]
  
\hline\rule{0pt}{11pt}
Random Forest \cite{ramos2020predicting} & Clinical variables & 0.65$\pm$0.02   & 0.43$\pm$0.01 &  0.65$\pm$0.02 & 0.64$\pm$0.02 & 0.64$\pm$0.02 & 0.63$\pm$0.02\\
3D-CNN \cite{Debs21} &  MRI modalities & 0.66$\pm$0.02   & 0.43$\pm$0.00 &    0.65$\pm$0.02 & 0.64$\pm$0.02 & 0.64$\pm$0.02 & 0.64$\pm$0.02\\
CNN-LSTM \cite{hatami2022EMBC} & MRI modalities & 0.62$\pm$0.04   &  0.42$\pm$0.01	&  0.63$\pm$0.04 &	0.61$\pm$0.04 &0.61$\pm$0.04& 0.60$\pm$0.03 \\
CNN-LSTM CL \cite{hatami2022ESOC} & MRI modalities & 0.68$\pm$0.05   &  0.38$\pm$0.01	&  0.69$\pm$0.02 &	0.65$\pm$0.03 &0.66$\pm$0.03& 0.64$\pm$0.02 \\
\bf{AE$^2$-LSTM}  & \bf{MRI modalities} & \bf{0.71$\pm$0.03}  & \bf{0.34$\pm$0.01} &  \bf{0.72$\pm$0.03} &	\bf{0.67$\pm$0.03} & \bf{0.67$\pm$0.03} & \bf{0.68$\pm$0.03}   \\
[2pt]
\hline
\end{tabular}
\end{tabular}
\end{center}
\label{table1}
\end{table*}

\subsection{Data Source and Preprocessing}

Patients were included from the HIBISCUS-STROKE cohort \cite{Debs21, hatami2022EMBC}, which is an ongoing monocentric observational cohort enrolling patients with an ischemic stroke due to a proximal intracranial artery occlusion treated by thrombectomy. 
In total 119 patients were analyzed. Inclusion criteria were: (1) patients with an anterior circulation stroke related to a proximal intracranial occlusion; (2) diffusion and perfusion MRI as baseline imaging; (3) patients treated by thrombectomy with or without intravenous thrombolysis. 

\begin{figure}
\centering
\centerline{\includegraphics[width=8.8cm]{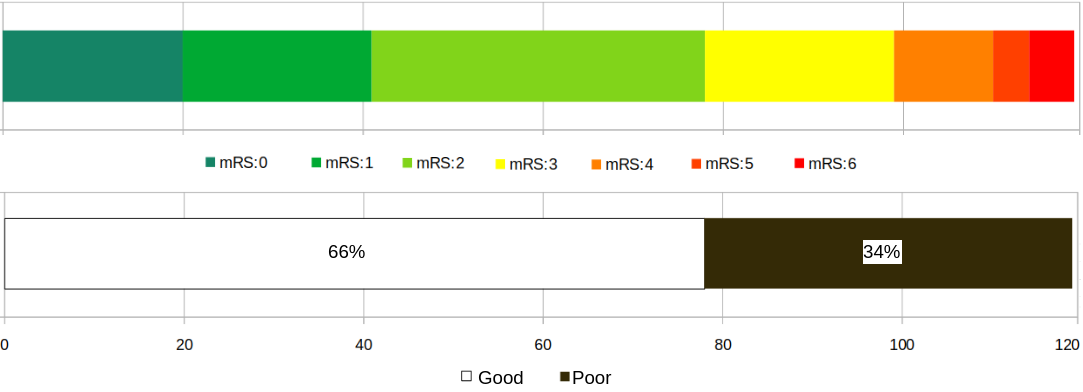}}
\vspace{-0.2cm}
\caption{Top) Distribution of the 3-month mRS scores, bottom) associated binarization into good \{0,1,2\} and poor outcome \{3,4,5,6\}. 66\% vs. 34\% good and poor outcomes respectively within our cohort.}
\label{fig_mrsdata}
\end{figure}

All patients underwent the following protocol (IRB number:00009118) at admission: DWI, and dynamic susceptibility-contrast perfusion imaging (DSC-PWI).
Final clinical outcome was assessed at 3-month during a face-to-face follow-up visit using the mRS. 
The distribution of the final mRS scores and its associated binarization into poor and good outcomes are shown in Figure \ref{fig_mrsdata}. In this paper, we used the binarized mRS for classifying patients' outcomes.


Parametric maps were extracted from the DSC-PWI by circular singular value decomposition of the tissue concentration curves (Olea Sphere, Olea Medical, France): CBF, CBV and Tmax. DSC-PWI parametric maps were coregistered within subjects to DWI using linear registration with Ants \cite{avants2011reproducible} and all MRI slices were of size $192 \times 192$. The skull from all patients was removed using FSL \cite{smith2001fsl}. Finally, images were normalized between 0 and 1 to ensure inter-patient standardization.

\subsection{Analysis and Results}
Six different measures are used to evaluate the performance of the models: classification accuracy (recognition rate), F1 score, sensitivity, specificity, Mean-Absolute-Error (MAE) and the Area Under the Curve (AUC). Ten independent runs with random seeds are performed, and means and standard deviations of the measures are reported. 

To compare our results, we proposed four baseline models. First baseline is a RF classifier inspired by Ramos et al. \cite{ramos2020predicting}. 
The inputs for the RF are the following clinical data: NIHSS baseline, age, door-to-puncture time and Fazekas scale. Second baseline is inspired by \cite{Debs21,zihni2020} where the input MR images are identical to our proposed model. It is an early fusion 3D-CNNs model, where the five MR modalities are concatenated and used as an unique input. 
In order to represent the architecture, we use $C_{(size)}$ and $F_{(size)}$ where $C$ is a 3D convolution ($3\times3\times3$) followed by a ReLU activation function, a batch-normalization and 3D max-pooling layers and $F$ is FC. Therefore, the CNN architecture is $C_{8}$-$C_{16}$-$C_{32}$-$C_{64}$-$C_{128}$-$C_{256}$-$F_{100}$. 
The third and fourth baselines are two similar CNN-LSTM models, one is trained in the standard Stochastic Gradient Descent (SGD) manner \cite{hatami2022EMBC} and the other using Curriculum Learning (CL) strategy \cite{hatami2022ESOC}. In SGD mini-batches are randomly sampled from the training set to feed into the learner to update its weights at each training iteration, while CL organizes the mini-batches according to an ascending level of difficulty.  

We carried out experiments in PyTorch. Five independent stacked AEs are trained, one for each MRI modality, to learn the module-specific features and another AE for learning multimodal features from the unimodal features. For simplicity, the architecture and parameters of the AEs are chosen similar and set as followings: \emph{Max Epochs:400}, \emph{L2 Weight Regularization:0.004}, \emph{Sparsity Regularization:4}, \emph{Sparsity Proportion:0.05}. In order to obtain the best results, different \emph{feature sizes:\{200,500,1000,2000\}} were tried.
The LSTM architecture consists of a sequence of input layers, two LSTM layers, a FC and a sigmoid layer with the half-mean-squared-error loss. In order to obtain the best results, Adam and SGD optimizers with learning rate of $1e-4$ were tried with a maximum number of epochs of $1000$ with early stopping and batch size of $32$. 
In our experiments, we explored different hidden nodes for LSTM $nh:\{50,100,500,1000\}$ and the best $nh:500$ was selected for optimal feature size of 1000 based on 5-fold cross-validation performance.

Table \ref{table1} reports the performance of the proposed AE$^2$-LSTM compared to different baselines and state-of-the-art models. The AE$^2$-LSTM model outperforms the other models in all of the six measures with AUC=0.71, MAE=0.34, Accuracy=72\%, Specificity=0.67, Sensitivity=0.67 and F1 score=0.68.

\section{Conclusions and Future Work}

A novel deep learning based architecture is proposed for predicting stroke patient outcome. The proposed AE$^2$-LSTM model outperformed the state-of-the-art models showing that it is a better fit for multimodal and volumetric data such as MRI. The AEs in this model carefully learn the multimodal features from different image modalities and the LSTM learns the sequential information in MRI sequences and map it to the final mRS score. 

As for future research, the proposed AE$^2$-LSTM model should be further validated on larger multi-center cohorts. Effects of different AE types and different imaging modalities should also be investigated. Adapting the model in order to use the clinical and biological variables together with the imaging modalities is another important future direction.  

\section{Acknowledgments}

This work was supported by the RHU MARVELOUS (ANR-16-RHUS-0009) of Universite Claude Bernard Lyon-1 (UCBL) and by the RHU BOOSTER (ANR-18-RHUS-0001), within the program "Investissements d'Avenir" operated by the French National Research Agency (ANR). 

\section{Compliance with Ethical Standards}
This study followed the principles of the Declaration of Helsinki, and was approved by the local ethics committee (IRB number: 00009118, September 8th, 2016). All subjects or their relatives signed an informed consent form.

\bibliographystyle{IEEEbib}
\bibliography{refs}
\end{document}